\pdfoutput=1

\documentclass[11pt]{article}

\usepackage[final]{acl}

\usepackage{times}
\usepackage{latexsym}
\usepackage{graphicx}
\usepackage{hyperref}
\usepackage[inline]{enumitem}
\usepackage{tcolorbox}

\usepackage[T1]{fontenc}

\usepackage[utf8]{inputenc}

\usepackage{microtype}

\usepackage{inconsolata}

\title{Facilitating Opinion Diversity through Hybrid NLP Approaches\\ \large Thesis Proposal}

\author{Michiel van der Meer \\
  LIACS \\
  Leiden University\\
  \texttt{m.t.van.der.meer@liacs.leidenuniv.nl} \\}

\begin{document}
\maketitle
\begin{abstract}

Modern democracies face a critical issue of declining citizen participation in decision-making.
Online discussion forums are an important avenue for enhancing citizen participation.
This thesis proposal
\begin{enumerate*}[label=(\arabic*)]
    \item identifies the challenges involved in facilitating large-scale online discussions with Natural Language Processing (NLP),
    \item suggests solutions to these challenges by incorporating hybrid human--AI technologies, and
    \item investigates what these technologies can reveal about individual perspectives in online discussions.
\end{enumerate*}
We propose a three-layered hierarchy for representing perspectives that can be obtained by a mixture of human intelligence and large language models. We illustrate how these representations can draw insights into the diversity of perspectives and allow us to investigate interactions in online discussions.
\end{abstract}

\section{Introduction}
Addressing societal problems, such as climate change, pandemics, and resource scarcity, requires citizen engagement. One way to enhance citizen participation is by engaging with the public directly in society-wide conversations on online platforms \citep{smith2009democratic, friess2015systematic}. Online discussions help identify the problem areas and possible solutions that fit the diverse needs of those affected \citep{surowiecki2004wisdom, dryzek2019crisis}.

Online discussions generate vast amounts of content, which is challenging to manage and navigate \citep{dahlberg2001internet}. Further, the content is scattered across time and threads, and it contains frequently repeating arguments and abundant unconnected ideas. This makes it difficult for users to know where to add new contributions, resulting in low-quality content \citep{klein2012enabling}. These issues can be addressed by employing moderators or facilitators, e.g., to structure the content of a discussion or to steer user interactions \citep{trenel2009facilitation}. However, given the amount of data, manually facilitating online discussions is not feasible.

Instead, we turn to NLP for interpreting text-based opinions at scale \citep{sun2017review}, powered by the recent surge of Large Language Models (LLMs) \citep{min2023recent, argyle2023out}. Central to our approach to facilitation is extracting structured \emph{perspectives} from users in a discussion. The perspectives provide high-level insights into the arguments employed by citizens \citep{vecchi2021towards} or the motivations underlying the opinions in a community \citep{weld2022makes}. These representations may, in turn, influence the facilitation strategies \citep{falk2021predicting} or shape policies following the discussion \citep{mouter2021public}.

Using NLP for analyzing opinions sourced from online platforms comes with its own set of challenges. For instance, online platforms have been centered on managing large volumes of information, e.g., through personalized recommendations \citep{adomavicius2005toward} or argument structuring \citep{iandoli2014socially} but have neglected inclusive design aspects \citep{shortall2022reason}. This can cause majority opinions to be heard while suppressing dissent voices \citep{neubaum2017monitoring}. Similarly, we see that LLMs capture majority opinions well, but do not distill all voices equally \citep[e.g.,][]{mustafaraj2011vocal, vandermeer2024empirical}. Further, LLMs lack deep social reasoning \citep{liang2021towards}, may be biased \citep{hartmann2023political, santurkar2023whose}, and make mistakes in ways humans cannot anticipate \citep{huang2023survey}. Finally, straightforward automated discussion analysis runs the danger of ignoring diverse opinions, which undermines the wisdom of the crowd effect \citep{lorenz2011social}. In light of these challenges, we ask our first research question:

\begin{description}[itemsep=-5pt, leftmargin=0em, topsep=0pt]
    \item[Q1] \emph{What fundamental issues arise in using NLP to analyze perspectives in online discussions?}
\end{description}

Next, our goal is to obtain structured information from online societal discussions that provide insights into the opinions involved. However, we see that NLP-based methods for analyzing online deliberation are limited in the degree to which \textbf{diverse} perspectives can be obtained. To combat these limitations, we develop an approach that adopts a ``hybrid'' mindset, i.e., incorporates humans-in-the-loop to address diversity directly. We leverage LLMs and humans jointly, with their different capacities for interpreting opinions from text. This leads to our second research question:

\begin{description}[itemsep=-5pt, leftmargin=0em, topsep=0pt]
    \item[Q2] \emph{How to combine human intelligence and NLP to effectively capture diverse perspectives?}
\end{description}

Finally, analyzing opinions, in practice, is modeled by different tasks. We propose a \textbf{perspective hierarchy} that incorporates \emph{stance, arguments, and personal values} to represent perspectives at different levels of abstraction. We base our model on the complementary skills of humans and NLP methods.
Higher-order abstractions, such as personal values, deeply motivate choices and the attitudes of individuals but are difficult to estimate automatically. Conversely, surface-level stance recognition tasks are more widely applicable but uncover little information about an individual's opinion. Each task has been investigated separately, but little is known about their interaction in online discussions. We, therefore, ask our third research question:

\begin{description}[itemsep=-5pt, leftmargin=0em, topsep=0pt]
    \item[Q3] \emph{How to combine different tasks for representing diverse opinions in online discussions?}
\end{description}

Sections~\ref{sec:fundamental},~\ref{sec:hybrid}, and~\ref{sec:values} describe our progress on the three questions. Section~\ref{sec:conclusions} concludes the paper.

\section{Use of NLP in Societal Discussions}
\label{sec:fundamental}

\begin{tcolorbox}[boxrule=0.25mm,left=2pt,right=2pt]
    \textbf{Q1} What fundamental issues arise in using NLP to analyze perspectives in online discussions?
\end{tcolorbox}

\noindent NLP research regarding the facilitation of online societal discussions has seen recent interest \citep[e.g.,][]{crossley2016combining, jelodar2020deep, xia2020exploring}.
Research is focused on \begin{enumerate*}[label=(\arabic*)]
    \item using NLP tools, in particular few-shot prompted LLMs, to analyze the discussions \citep[e.g.,][]{xia2020exploring, syed2023indicative}, and
    \item using discussion data to benchmark the capabilities of NLP tools \citep[e.g.,][]{feng2023pretraining}.
\end{enumerate*}
In the next two sections, we outline related work in these directions, highlighting fundamental issues that cross-cut techniques and applications.

\subsection{Discussion Analysis}
\label{sec:discussion-analysis}
Online social interaction through text is common, and the use of NLP for analyzing large amounts of such data is mainstream \citep{liu2012sentiment}. Discussions happen in various specific contexts, e.g., reviews~\citep{jo2011aspect} or e-learning~\citep{davies2005performance}, but also broader contemporary topics such as climate change~\citep{lorcher2017discussing}. Their scale, combined with their pertinence makes analyzing such discussions interesting.

Analyzing how humans express themselves through text is the core task in many NLP areas, e.g., Opinion Summarization \citep{liu2012sentiment}, Argument Mining \citep{lawrence2020argument}, Sentiment Analysis \citep{wankhade2022survey}, and Value Classification \citep{hoover2020moral}. These tasks lie at the heart of creating insights into online (political) discourse and may be used e.g. for estimating the quality of discussions \citep{steenbergen2003measuring}, extracting the arguments involved \citep{lapesa2023mining}, or reasoning over inconsistencies between choices and their justifications \citep{liscio2024value}. In the age of LLMs, these tasks have seen considerable performance improvements \citep{jiang2023mistral}, though new challenges such as dealing with shortcut learning \citep{geirhos2020shortcut} or mitigating social biases \citep{liang2021towards} arise.

Extracting diverse views from online discussions is challenging for three reasons.
First, data sourced from social media platforms inherits biases that are present on these platforms, including fake news, trolling, and polarization \citep{cinelli2021echo}. This impacts how opinions are shaped \citep{hocevar2014social} and the distribution of opinions \citep{xiong2014opinion}.
Second, when analyzing the opinions about societal issues, it is necessary to realize that not all citizens have equal access due to the digital divide \citep{cullen2001addressing} or differences in tech-illiteracy \citep{knobel2008digital}. This makes the users in online discussions biased and less diverse.
Third, since users are free to join in discussions of their choosing, there may be undesired echo chambers or self-selection effects among the messages seen by users \citep{song2020dynamics}.

Despite these challenges, we can use NLP to investigate questions about human behavior at scale \citep{lazer2009computational}. Analyses about behavior may lead to insights on both individual and group levels. This can be useful for improving democratic processes \citep{collins2019examining}, but also applies in other areas, such as faithfully interpreting product feedback \citep{bar2021every}, service improvement \citep{skiera2022using}, or course management \citep{lin2009discovering}.

\subsection{Benchmarking}
We can employ discussion analysis to benchmark how well NLP approaches understand opinionated text. In benchmarking, we test the analysis procedure, and models used, for possible mistakes and biases. Representing subjectivity is difficult since LLMs do not faithfully capture the full range of opinions \citep{durmus2023towards, hendrycks2020aligning, vandermeer2024empirical}. Whether LLMs can learn to represent them in the future remains unclear \citep{wei2022emergent, schaeffer2024emergent}, but research suggests that they cannot \citep{feng2023pretraining, argyle2023out}, in part due to the limitations mentioned in Section~\ref{sec:discussion-analysis}. Therefore, we work with the assumption that this is a fundamental limitation of LLMs, and we have to find other approaches to improving diversity.\footnote{Although linguistic diversity generally refers to diversity of language proficiencies \citep{joshi2020state, dingemanseliesenfeld2022text}, we are specifically interested in diversity in arguments, communication styles, and values in online discussions.}

Creating diversity-enhancing techniques is gaining traction in NLP, but there are several aspects of diversity. For instance, creating more diverse news recommender systems is a common goal \citep{laban2022discord, wu2020sentirec} for shaping an individual's perspective \citep{bakshy2015exposure}. Others strive to make LLMs better represent a diverse group of annotators based on their labeling behavior and demographics \citep{bakker2022fine, lahoti2023improving}.
In such approaches, models have a large reliance on annotated data. Labels are obtained from a few human annotators per instance, and often aggregated by majority voting, painting an incomplete picture of the true range of interpretations for a potentially controversial text \citep{plank2022problem}.
The role of subjectivity in these tasks remains unclear \citep{aroyo2015truth, cabitza2023toward}. This holds for traditional supervised learning, but also for the latest trends in instruction-tuning \citep{uma2021learning, wang2024far}.

In the rest of this proposal, we argue that the aforementioned challenges can be overcome by using LLMs to \textbf{assist humans} in mining opinionated text data rather than replacing them, and we provide an example of how hybrid approaches can uncover perspectives of the opinion holders.

\section{Hybrid Intelligence}
\label{sec:hybrid}

\begin{figure}[tb]
    \centering
    \includegraphics[width=\columnwidth]{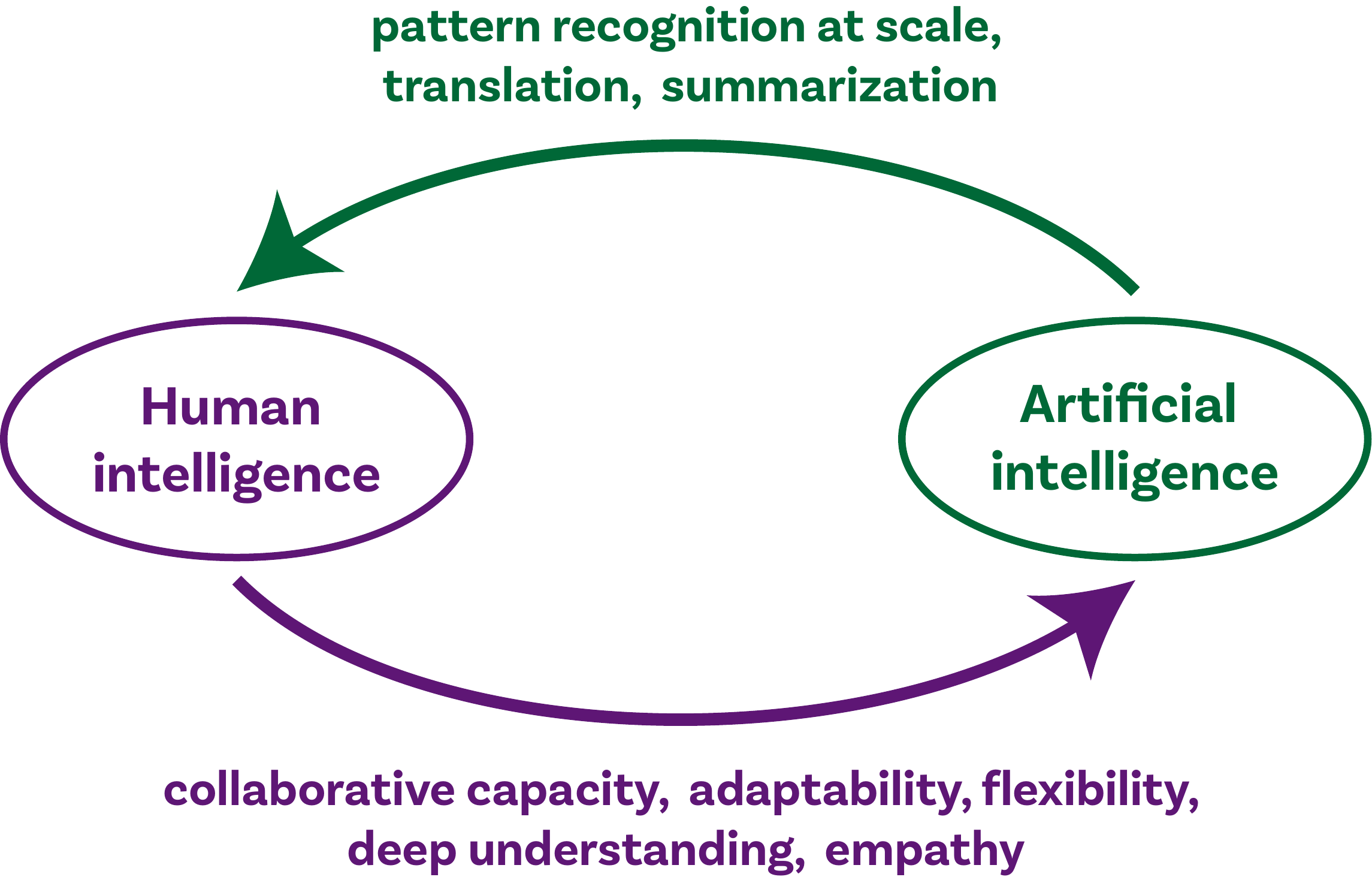}
    \caption{Feedback loops in Hybrid Intelligence.}
    \label{fig:hybrid}
\end{figure}

\begin{tcolorbox}[boxrule=0.25mm,left=2pt,right=2pt]
    \textbf{Q2} How to combine human intelligence and NLP to effectively capture diverse perspectives?
\end{tcolorbox}
\noindent Central to our proposal on facilitating deliberation is the notion of \emph{hybrid intelligence} \citep{dellermann2019hybrid, akata2020research, DellAnna-2024-AAMAS-HIQuality}. In Hybrid Intelligent Systems (HISs), artificial intelligence is a collaborator that enhances human abilities such as reasoning, decision-making, and problem-solving \citep{tiddi2023knowledge}. Hybrid intelligence aims to augment intellect, creating a synergy between humans and NLP. For supporting online discussions, we combine the strengths of human intelligence with LLMs, highlighting bidirectional gains, as shown in Figure~\ref{fig:hybrid}.

\subsection{Related Work}
NLP has had a profound impact on how researchers analyze human behavior at scale. To do so responsibly, we must ensure that these methods do so effectively while upholding democratic values.
Previous work on hybrid approaches for NLP includes user adaptation \citep{lynn2017human}, human-in-the-loop computing \citep{wang2021putting}, human-AI interaction \citep{heer2019agency} and others \citep[e.g.,][]{ding2023harnessing, costa2022no}. Recent interest in explainable AI has focused on human understanding of NLP models \citep{lertvittayakumjorn2021explanation}. Specifically for NLP, much focus is on approaches that mix crowd, expert, and automated decision-making, which have been applied to analyzing discussion content \citep{kong2022slipping, pacheco2023interactive}.
However, these approaches have a one-way interaction between the NLP model and humans, as we will describe in the next section.

\subsection{Approach}
We observe that LLMs still have many challenges to overcome in representing diverse perspectives (Section~\ref{sec:fundamental}). Discussions are deeply human, who can adapt to incomplete and informal argumentation, behave flexibly, and provide empathic responses to foster collaboration. Thus, humans and NLP can benefit from each other. In the next paragraphs, we examine each benefit in either direction (humans aiding NLP or NLP aiding humans) separately, and lastly illustrate how both can be incorporated into an overall hybrid method.

\paragraph{Humans aiding NLP}
Humans provide the data that the NLP tools perform their analysis on, as gathered from interactions between different stakeholders, including casual and power users, moderators, or even site admins \citep{saxena2022users}. They provide text and behavioral data, such as post-voting, which we in turn can use to analyze their attitude.
Furthermore, NLP approaches learn from labeled data, obtained from annotators who observe a given text and draw labels from a predefined set of classes. Much room for making these procedures more informative exist, such as expanding the label set \citep{van2022three}, including free-form text response \citep{ouyang2023shifted}, asking a crowd of annotators rather than individuals \citep{nie2020learn}, and more \citep[e.g.,][]{plank2022problem, santy2023nlpositionality}.


\paragraph{NLP aiding humans}
NLP aids humans in online discussions in multiple ways. While we have mostly discussed the analysis of large-scale discussion data, there is a broader potential impact of NLP technologies in online deliberations \citep{tomavsev2020ai}. First, NLP may enable, rather than restrict, access to certain services, for example by using automatic translation to account for different language proficiencies. Second, since humans suffer from cognitive biases, NLP models may offer an alternative interpretation of the content. Machines do not get bored and consider each sample identically. Third, NLP models mirror biases captured in the data, which allows for obtaining synthetic opinion data or exposing biases in discussions. Lastly, since their scale, speed, and accessibility to researchers are advancing quickly, we can experiment with them rapidly.

\paragraph{Combination}
Existing work mostly offers one-directional benefits, either machine- or human-oriented. We see that NLP methods are biased, leading to questions about the soundness of the analysis. Humans can repair biases and provide deeper interpretations, contexts, and explanations. Furthermore, we see that there are many opportunities for NLP to aid humans. Completing the loop allows bootstrapping: traversing the two feedback loops shown in Fig.~\ref{fig:hybrid}, iteratively refining the analysis procedure while performing discussion analyses. By building on the bidirectional contributions, we allow for continual improvement.

Our work involves discussion analysis approaches that involve \begin{enumerate*}[label=(\arabic*)]
    \item selecting samples for human inspection that are interesting to annotate,
    \item accounting for diversity (e.g., leveraging contextualized embeddings \citep{reimers2019sentence}),
    \item seeking labels from multiple annotators.
\end{enumerate*} We find that a hybrid approach can capture more diverse interpretations of the arguments in a discussion than a purely manual or purely automatic approach \citep{vandermeer2022hyena, van2024hybrid}. When extracting arguments from online comments, human annotators are more precise than NLP methods. At the same time, we use sampling based on the maximum embedding distance to ensure diverse content is observed \citep{basu2004active} and automatically merge similar arguments \citep{chai2016cost}. In this setup, we obtain labels from a crowd over diverse samples that promote perspective-taking. After the annotation, our method outputs a summary of the high-level argument involved, while annotators were able to develop their understanding of controversial discussions. Moreover, we can also actively diversify which annotator we query an annotation from. We observe that an active selection of diverse annotators can inform a model more quickly of the label distribution underlying subjective tasks in cases where the annotator pool is large \citep{van2024annotator}.

Developing hybrid approaches requires a new evaluation paradigm. We need to compare our method's effectiveness with human-only and machine-only baselines. In NLP, test sets are usually collected manually. This may make the upper bound on performance unfair, though performance gaps between hybrid and manual approaches can be addressed \citep{xu2021humanly, fluri2023evaluating}.

\section{Perspective Hierarchy}
\label{sec:values}

\begin{tcolorbox}[boxrule=0.25mm,left=2pt,right=2pt]
    \textbf{Q3} How to combine different tasks for representing diverse opinions in online discussions?
\end{tcolorbox}

\noindent Given that NLP can process large amounts of discussion data, but is limited in its capabilities (Section~\ref{sec:fundamental}), and that we may construct hybrid procedures to account for these limits (Section~\ref{sec:hybrid}), we address the challenge on how to capture perspectives. Uncovering them from online societal discussions requires a representation for identifying how people feel about potential decisions, how this is communicated in the discussions, and what their underlying motivations are.

\begin{figure}[tb]
    \centering
    \includegraphics[width=.84\columnwidth]{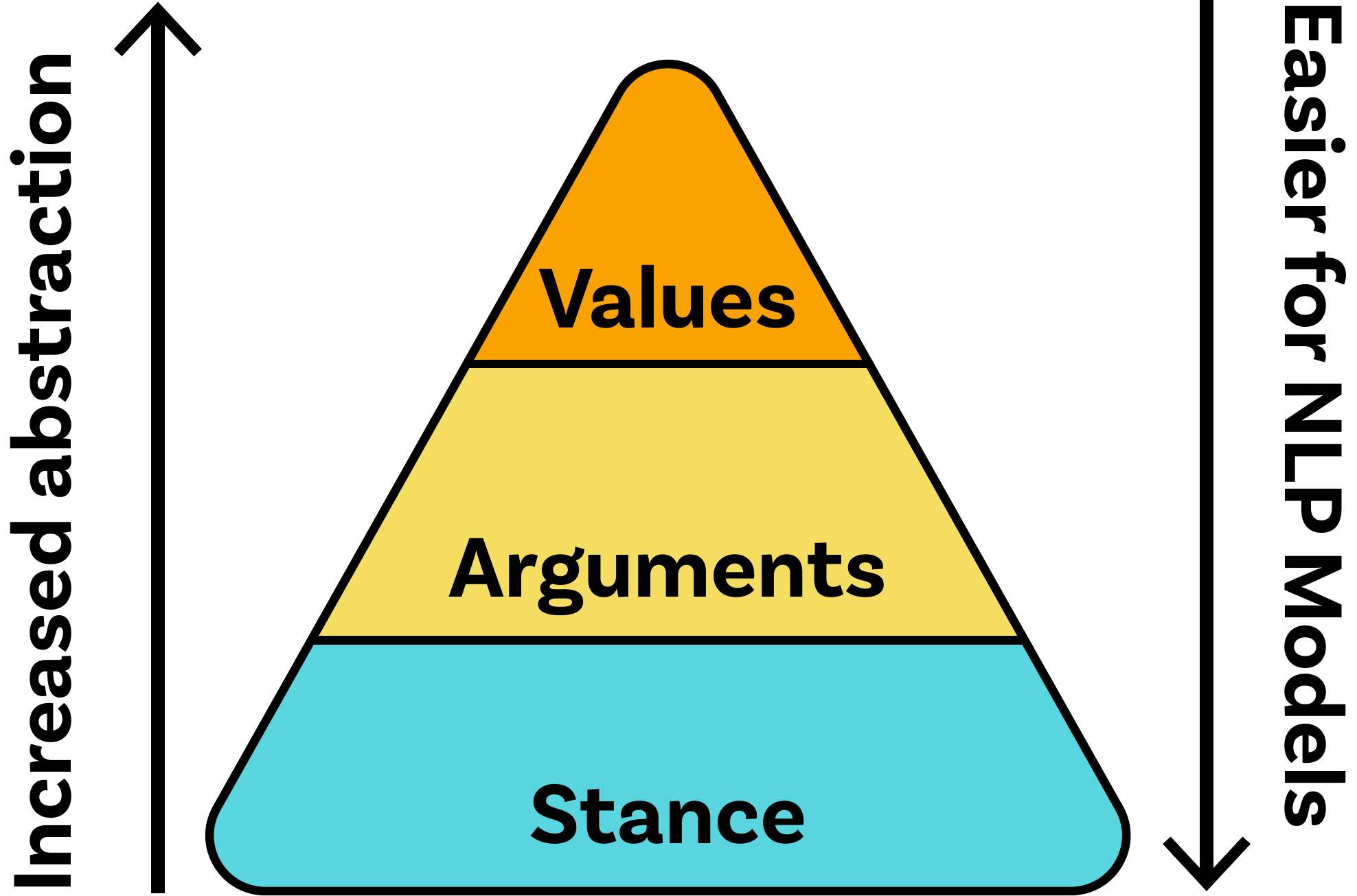}
    \caption{The perspective hierarchy. The higher the level of abstraction, the more human intelligence is required for interpreting the component.}
    \label{fig:hierarch}
\end{figure}

\subsection{Related Work}
Few attempts to represent perspectives holistically exist \citep{chen2019seeing, vanson2016grasp}. These works focus on annotating utterances for low-level claim information \citep{morante2020annotating}, or investigating some of the reasoning behind the views held in discussions \citep{draws2022comprehensive}. Stances and arguments are inherently linked in argumentation models \citep{toulmin2003uses, van2015argumentation}, and form the basis of frameworks for representing perspectives \citep{wiebe2005annotating, chen2022design}.

However, neither stance nor arguments aim to represent opinions on a deeper personal level. A fundamental concept for explaining the motivations underlying opinions and actions is personal values \citep{schwartz2012overview}. There are various theories of personal values \citep[e.g.,][]{rokeach1967rokeach, schwartz2012overview, graham2013moral}. Preferences among values describe the attitude of individuals and groups \citep{ponizovskiy2020development}, and can be extracted from behavioral cues to investigate political affiliation \citep{roy2021identifying}, perform moral reasoning \citep{mooijman2018moralization}, or positively influence lifestyle \citep{de2023contextual}.
Values are abstract and need to be interpreted inside their context, making it difficult for both humans and NLP methods to reliably measure them \citep{liscio2023value}. One way to contextualize them is to connect values to argumentation, focusing on how choices are justified and reasoned over \citep{kiesel2022identifying}. Using this insight, we incorporate personal values into our perspective representation and aim to obtain them using a hybrid approach.

\subsection{Approach}
We propose a perspective hierarchy to represent a person's perspective at different levels of abstraction, shown in Figure~\ref{fig:hierarch}. Our perspective hierarchy is composed of stances, arguments, and values.

\begin{description}[itemsep=-5pt, leftmargin=0pt]
    \item[Stance] Whether, or how much, support or opposition is expressed to a claim. Stance detection has been studied extensively and remains a popular task for investigating opinions on claims \citep{kuccuk2020stance}.
    \item[Arguments] The reasons given for adopting a stance towards a claim. In real-world contexts, argumentation manifests in many forms and is predominantly informal \citep{sep-logic-informal}. Mining arguments from text works well within known contexts \citep{ein2020corpus}, but suffers from implicit reasoning \citep{habernal2018argument}. Hence, we require more human guidance to correct for possible mistakes in automated methods.
    \item[Values] The motivations underlying opinions and actions \citep{schwartz2012overview}. Values are communicated implicitly through actions or written motivations. Estimating values automatically remains difficult even within their context \citep{kiesel2023semeval}. Only through iterative hybrid procedures can we accurately reason about preferences among human values.
\end{description}

\paragraph{Mining Perspective Hierarchies} We illustrate how we used data from large online social media platforms to investigate perspective hierarchies for individuals \citep{vandermeer2023differences}. Our main objective is to investigate whether we can connect stances and values directly, omitting arguments, to challenge their inclusion in the hierarchy.

Given a societal discussion on an online platform \citep{pougue2021debagreement}, we first identify relevant controversial topics and apply our automated methods for obtaining stances and value preferences. Because of the aforementioned limitations, we utilize the human-in-the-loop approach to uncover possible mistakes from the extraction pipeline. In particular, we compare human-provided self-reported value preferences to those estimated from behavioral data. Using this data, we can
\begin{enumerate*}[label=(\arabic*)]
    \item compare how well the automated approaches work versus manual ones,
    \item mix information from self-reported and behavior-based value preferences, and
    \item investigate the relationship between components of the perspective hierarchy to answer questions about human behavior.
\end{enumerate*}

We probed the relationship between disagreements in stance and deeper conflicts in values. Our experiments show that when values are diverse, conflicts in values can correlate to stance disagreement. Based on purely automated estimations, this evidence is weak. When we incorporate human-provided self-reports, the evidence becomes stronger, showing that the hybrid approach is crucial to performing a meaningful analysis. On the other hand, when strong value diversity is absent, we cannot correlate disagreement and value conflict directly. Thus, we require a more complete picture, and should therefore incorporate the arguments to complete the perspective hierarchy.

\section{Conclusions}
\label{sec:conclusions}
We identified the strengths and weaknesses of using NLP to represent diverse perspectives in online societal discussions.
NLP techniques, in particular few-shot prompting with LLMs, allow us to analyze discussion data for perspectives at a large scale.
However, open challenges include
\begin{enumerate*}[label=(\arabic*)]
    \item a difficulty in acquiring opinions from diverse opinion holders, and
    \item limitations of LLMs to represent minority opinions.
\end{enumerate*}
Our approach combines the complementary abilities of humans and LLMs into hybrid intelligence methods to obtain better analyses than automated or manual analysis alone. We propose a perspective hierarchy to guide the investigation of human behavior in online societal discussions at scale. We find that this hierarchy is useful for uncovering perspectives, for instance, in observing that diversity in opinions can be signaled by differences among value preferences.

\subsection*{Future Directions}
\label{sec:future}

First, integrating human and artificial work requires careful task balancing. In some cases, obtaining an automated judgment from an LLM is sufficient, but in others, we need to query a pool of diverse human annotators. We can use frameworks like learning to defer \citep{madras2018predict} or other active learning approaches \citep{baumler2023examples} to directly obtain diverse opinions \citep{waterschoot2022detecting}.

Second, 
evaluation of hybrid intelligence systems requires novel benchmarking paradigms. Existing benchmarks are usually annotated manually and composed out of many individual existing datasets, and therefore lack a faithful representation of the dynamic context of real-world applications \citep{chang2024survey}.
Alternative approaches can instead incorporate interactive crowd-sourced benchmarks that develop over time \citep{kiela-etal-2021-dynabench}, or turn to use-case-specific evaluation, leveraging objective behavioral cues to assess our methods, e.g., in measuring interaction structure to reveal the quality of a conversation \citep{santamaria2022evaluating}.

Lastly, our proposed hybrid human-AI approach engages with citizens to learn their perspectives. We represent the cares, incentives, and preferences of those involved in societal discussions. 
In the long run, we may be able to adopt components in the perspective hierarchy for not only facilitating discussions but supporting negotiations \citep{renting2022automated} among societal stakeholders, e.g., on which portfolio of choices to make to combat a pandemic \citep{mouter2021public}.

\bibliography{custom}

\begin{thebibliography}{126}
\expandafter\ifx\csname natexlab\endcsname\relax\def\natexlab#1{#1}\fi

\bibitem[{Adomavicius and Tuzhilin(2005)}]{adomavicius2005toward}
G.~Adomavicius and A.~Tuzhilin. 2005.
\newblock \href {https://doi.org/10.1109/TKDE.2005.99} {Toward the next
  generation of recommender systems: a survey of the state-of-the-art and
  possible extensions}.
\newblock \emph{IEEE Transactions on Knowledge and Data Engineering},
  17(6):734--749.

\bibitem[{Akata et~al.(2020)Akata, Balliet, de~Rijke, Dignum, Dignum, Eiben,
  Fokkens, Grossi, Hindriks, Hoos, Hung, Jonker, Monz, Neerincx, Oliehoek,
  Prakken, Schlobach, van~der Gaag, van Harmelen, van Hoof, van Riemsdijk, van
  Wynsberghe, Verbrugge, Verheij, Vossen, and Welling}]{akata2020research}
Zeynep Akata, Dan Balliet, Maarten de~Rijke, Frank Dignum, Virginia Dignum,
  Guszti Eiben, Antske Fokkens, Davide Grossi, Koen Hindriks, Holger Hoos,
  Hayley Hung, Catholijn Jonker, Christof Monz, Mark Neerincx, Frans Oliehoek,
  Henry Prakken, Stefan Schlobach, Linda van~der Gaag, Frank van Harmelen,
  Herke van Hoof, Birna van Riemsdijk, Aimee van Wynsberghe, Rineke Verbrugge,
  Bart Verheij, Piek Vossen, and Max Welling. 2020.
\newblock \href {https://doi.org/10.1109/MC.2020.2996587} {A research agenda
  for hybrid intelligence: Augmenting human intellect with collaborative,
  adaptive, responsible, and explainable artificial intelligence}.
\newblock \emph{Computer}, 53(8):18--28.

\bibitem[{Argyle et~al.(2023)Argyle, Busby, Fulda, Gubler, Rytting, and
  Wingate}]{argyle2023out}
Lisa~P. Argyle, Ethan~C. Busby, Nancy Fulda, Joshua~R. Gubler, Christopher
  Rytting, and David Wingate. 2023.
\newblock \href {https://doi.org/10.1017/pan.2023.2} {Out of one, many: Using
  language models to simulate human samples}.
\newblock \emph{Political Analysis}, 31(3):337–351.

\bibitem[{Aroyo and Welty(2015)}]{aroyo2015truth}
Lora Aroyo and Chris Welty. 2015.
\newblock \href {https://doi.org/10.1609/aimag.v36i1.2564} {Truth is a lie:
  Crowd truth and the seven myths of human annotation}.
\newblock \emph{AI Magazine}, 36(1):15--24.

\bibitem[{Bakker et~al.(2022)Bakker, Chadwick, Sheahan, Tessler,
  Campbell-Gillingham, Balaguer, McAleese, Glaese, Aslanides, Botvinick, and
  Summerfield}]{bakker2022fine}
Michiel Bakker, Martin Chadwick, Hannah Sheahan, Michael Tessler, Lucy
  Campbell-Gillingham, Jan Balaguer, Nat McAleese, Amelia Glaese, John
  Aslanides, Matt Botvinick, and Christopher Summerfield. 2022.
\newblock \href
  {https://proceedings.neurips.cc/paper_files/paper/2022/file/f978c8f3b5f399cae464e85f72e28503-Paper-Conference.pdf}
  {Fine-tuning language models to find agreement among humans with diverse
  preferences}.
\newblock In \emph{Advances in Neural Information Processing Systems},
  volume~35, pages 38176--38189. Curran Associates, Inc.

\bibitem[{Bakshy et~al.(2015)Bakshy, Messing, and Adamic}]{bakshy2015exposure}
Eytan Bakshy, Solomon Messing, and Lada~A. Adamic. 2015.
\newblock \href {https://doi.org/10.1126/science.aaa1160} {Exposure to
  ideologically diverse news and opinion on facebook}.
\newblock \emph{Science}, 348(6239):1130--1132.

\bibitem[{Bar-Haim et~al.(2021)Bar-Haim, Eden, Kantor, Friedman, and
  Slonim}]{bar2021every}
Roy Bar-Haim, Lilach Eden, Yoav Kantor, Roni Friedman, and Noam Slonim. 2021.
\newblock \href {https://doi.org/10.18653/v1/2021.acl-long.262} {Every bite is
  an experience: {K}ey {P}oint {A}nalysis of business reviews}.
\newblock In \emph{Proceedings of the 59th Annual Meeting of the Association
  for Computational Linguistics and the 11th International Joint Conference on
  Natural Language Processing (Volume 1: Long Papers)}, pages 3376--3386,
  Online. Association for Computational Linguistics.

\bibitem[{Basu et~al.(2004)Basu, Banerjee, and Mooney}]{basu2004active}
Sugato Basu, Arindam Banerjee, and Raymond~J. Mooney. 2004.
\newblock \href {https://doi.org/10.1137/1.9781611972740.31} {Active
  semi-supervision for pairwise constrained clustering}.
\newblock In \emph{Proceedings of the 2004 SIAM International Conference on
  Data Mining (SDM)}, pages 333--344.

\bibitem[{Baumler et~al.(2023)Baumler, Sotnikova, and
  Daum{\'e}~III}]{baumler2023examples}
Connor Baumler, Anna Sotnikova, and Hal Daum{\'e}~III. 2023.
\newblock \href {https://doi.org/10.18653/v1/2023.findings-acl.658} {Which
  examples should be multiply annotated? active learning when annotators may
  disagree}.
\newblock In \emph{Findings of the Association for Computational Linguistics:
  ACL 2023}, pages 10352--10371, Toronto, Canada. Association for Computational
  Linguistics.

\bibitem[{Cabitza et~al.(2023)Cabitza, Campagner, and
  Basile}]{cabitza2023toward}
Federico Cabitza, Andrea Campagner, and Valerio Basile. 2023.
\newblock \href {https://doi.org/10.1609/aaai.v37i6.25840} {Toward a
  perspectivist turn in ground truthing for predictive computing}.
\newblock \emph{Proceedings of the AAAI Conference on Artificial Intelligence},
  37(6):6860--6868.

\bibitem[{Chai et~al.(2016)Chai, Li, Li, Deng, and Feng}]{chai2016cost}
Chengliang Chai, Guoliang Li, Jian Li, Dong Deng, and Jianhua Feng. 2016.
\newblock \href {https://doi.org/10.1145/2882903.2915252} {Cost-effective
  crowdsourced entity resolution: A partial-order approach}.
\newblock In \emph{Proceedings of the 2016 International Conference on
  Management of Data}, SIGMOD '16, page 969–984, New York, NY, USA.
  Association for Computing Machinery.

\bibitem[{Chang et~al.(2024)Chang, Wang, Wang, Wu, Yang, Zhu, Chen, Yi, Wang,
  Wang, Ye, Zhang, Chang, Yu, Yang, and Xie}]{chang2024survey}
Yupeng Chang, Xu~Wang, Jindong Wang, Yuan Wu, Linyi Yang, Kaijie Zhu, Hao Chen,
  Xiaoyuan Yi, Cunxiang Wang, Yidong Wang, Wei Ye, Yue Zhang, Yi~Chang,
  Philip~S. Yu, Qiang Yang, and Xing Xie. 2024.
\newblock \href {https://doi.org/10.1145/3641289} {A survey on evaluation of
  large language models}.
\newblock \emph{ACM Trans. Intell. Syst. Technol.}, 15(3).

\bibitem[{Chen et~al.(2019)Chen, Khashabi, Yin, Callison-Burch, and
  Roth}]{chen2019seeing}
Sihao Chen, Daniel Khashabi, Wenpeng Yin, Chris Callison-Burch, and Dan Roth.
  2019.
\newblock \href {https://doi.org/10.18653/v1/N19-1053} {Seeing things from a
  different angle:discovering diverse perspectives about claims}.
\newblock In \emph{Proceedings of the 2019 Conference of the North {A}merican
  Chapter of the Association for Computational Linguistics: Human Language
  Technologies, Volume 1 (Long and Short Papers)}, pages 542--557, Minneapolis,
  Minnesota. Association for Computational Linguistics.

\bibitem[{Chen et~al.(2022)Chen, Liu, Uyttendaele, Zhang, Bruno, and
  Roth}]{chen2022design}
Sihao Chen, Siyi Liu, Xander Uyttendaele, Yi~Zhang, William Bruno, and Dan
  Roth. 2022.
\newblock \href {https://doi.org/10.18653/v1/2022.findings-naacl.22} {Design
  challenges for a multi-perspective search engine}.
\newblock In \emph{Findings of the Association for Computational Linguistics:
  NAACL 2022}, pages 293--303, Seattle, United States. Association for
  Computational Linguistics.

\bibitem[{Cinelli et~al.(2021)Cinelli, Morales, Galeazzi, Quattrociocchi, and
  Starnini}]{cinelli2021echo}
Matteo Cinelli, Gianmarco De~Francisci Morales, Alessandro Galeazzi, Walter
  Quattrociocchi, and Michele Starnini. 2021.
\newblock \href {https://doi.org/10.1073/pnas.2023301118} {The echo chamber
  effect on social media}.
\newblock \emph{Proceedings of the National Academy of Sciences},
  118(9):e2023301118.

\bibitem[{Collins and Nerlich(2019)}]{collins2019examining}
Luke Collins and Brigitte Nerlich. 2019.
\newblock Examining user comments for deliberative democracy: A corpus-driven
  analysis of the climate change debate online.
\newblock In \emph{Climate Change Communication and the Internet}, pages
  41--59. Routledge.

\bibitem[{Crossley et~al.(2016)Crossley, Paquette, Dascalu, McNamara, and
  Baker}]{crossley2016combining}
Scott Crossley, Luc Paquette, Mihai Dascalu, Danielle~S. McNamara, and Ryan~S.
  Baker. 2016.
\newblock \href {https://doi.org/10.1145/2883851.2883931} {Combining
  click-stream data with nlp tools to better understand mooc completion}.
\newblock In \emph{Proceedings of the Sixth International Conference on
  Learning Analytics \& Knowledge}, LAK '16, page 6–14, New York, NY, USA.
  Association for Computing Machinery.

\bibitem[{Cullen(2001)}]{cullen2001addressing}
Rowena Cullen. 2001.
\newblock \href {https://doi.org/10.1108/14684520110410517} {Addressing the
  digital divide}.
\newblock \emph{Online information review}, 25(5):311--320.

\bibitem[{Dahlberg(2001)}]{dahlberg2001internet}
Lincoln Dahlberg. 2001.
\newblock \href {https://doi.org/10.1080/13691180110097030} {The internet and
  democratic discourse: Exploring the prospects of online deliberative forums
  extending the public sphere}.
\newblock \emph{Information, communication \& society}, 4(4):615--633.

\bibitem[{Davies and Graff(2005)}]{davies2005performance}
Jo~Davies and Martin Graff. 2005.
\newblock \href
  {https://doi.org/https://doi.org/10.1111/j.1467-8535.2005.00542.x}
  {Performance in e-learning: online participation and student grades}.
\newblock \emph{British Journal of Educational Technology}, 36(4):657--663.

\bibitem[{de~Boer et~al.(2023)de~Boer, van~der Waa, van Gent, Smit, Korteling,
  van Stokkum, and Neerincx}]{de2023contextual}
Maaike~H de~Boer, Jasper van~der Waa, Sophie van Gent, Quirine~T.S. Smit,
  Wouter Korteling, Robin~M. van Stokkum, and Mark Neerincx. 2023.
\newblock \href {https://mrc.kriwi.de/2023/download/paper-1-1.pdf} {A
  contextual hybrid intelligent system design for diabetes lifestyle
  management}.
\newblock In \emph{International Workshop Modelling and Representing Context,
  ECAI}, volume~23.

\bibitem[{Dell'Anna et~al.(2024)Dell'Anna, Murukannaiah, Dudzik, Grossi,
  Jonker, Oertel, and Yolum}]{DellAnna-2024-AAMAS-HIQuality}
Davide Dell'Anna, Pradeep~K. Murukannaiah, Bernd Dudzik, Davide Grossi,
  Catholijn~M. Jonker, Catharine Oertel, and P{ı}nar Yolum. 2024.
\newblock Toward a quality model for hybrid intelligence teams.
\newblock In \emph{Proceedings of the 23rd International Conference on
  Autonomous Agents and Multiagent Systems}, pages 1--10, Auckland.
\newblock To appear.

\bibitem[{Dellermann et~al.(2019)Dellermann, Ebel, S{\"o}llner, and
  Leimeister}]{dellermann2019hybrid}
Dominik Dellermann, Philipp Ebel, Matthias S{\"o}llner, and Jan~Marco
  Leimeister. 2019.
\newblock \href {https://doi.org/10.1007/s12599-019-00595-2} {Hybrid
  intelligence}.
\newblock \emph{Business \& Information Systems Engineering}, 61:637--643.

\bibitem[{Ding et~al.(2023)Ding, Smith-Renner, Zhang, Tetreault, and
  Jaimes}]{ding2023harnessing}
Zijian Ding, Alison Smith-Renner, Wenjuan Zhang, Joel Tetreault, and Alejandro
  Jaimes. 2023.
\newblock \href {https://doi.org/10.18653/v1/2023.findings-emnlp.217}
  {Harnessing the power of {LLM}s: Evaluating human-{AI} text co-creation
  through the lens of news headline generation}.
\newblock In \emph{Findings of the Association for Computational Linguistics:
  EMNLP 2023}, pages 3321--3339, Singapore. Association for Computational
  Linguistics.

\bibitem[{Dingemanse and Liesenfeld(2022)}]{dingemanseliesenfeld2022text}
Mark Dingemanse and Andreas Liesenfeld. 2022.
\newblock \href {https://doi.org/10.18653/v1/2022.acl-long.385} {From text to
  talk: {H}arnessing conversational corpora for humane and diversity-aware
  language technology}.
\newblock In \emph{Proceedings of the 60th Annual Meeting of the Association
  for Computational Linguistics (Volume 1: Long Papers)}, pages 5614--5633,
  Dublin, Ireland. Association for Computational Linguistics.

\bibitem[{Draws et~al.(2022)Draws, Inel, Tintarev, Baden, and
  Timmermans}]{draws2022comprehensive}
Tim Draws, Oana Inel, Nava Tintarev, Christian Baden, and Benjamin Timmermans.
  2022.
\newblock \href {https://doi.org/10.1145/3498366.3505812} {Comprehensive
  viewpoint representations for a deeper understanding of user interactions
  with debated topics}.
\newblock In \emph{Proceedings of the 2022 Conference on Human Information
  Interaction and Retrieval}, CHIIR '22, page 135–145, New York, NY, USA.
  Association for Computing Machinery.

\bibitem[{Dryzek et~al.(2019)Dryzek, Bächtiger, Chambers, Cohen, Druckman,
  Felicetti, Fishkin, Farrell, Fung, Gutmann, Landemore, Mansbridge, Marien,
  Neblo, Niemeyer, Setälä, Slothuus, Suiter, Thompson, and
  Warren}]{dryzek2019crisis}
John~S. Dryzek, André Bächtiger, Simone Chambers, Joshua Cohen, James~N.
  Druckman, Andrea Felicetti, James~S. Fishkin, David~M. Farrell, Archon Fung,
  Amy Gutmann, Hélène Landemore, Jane Mansbridge, Sofie Marien, Michael~A.
  Neblo, Simon Niemeyer, Maija Setälä, Rune Slothuus, Jane Suiter, Dennis
  Thompson, and Mark~E. Warren. 2019.
\newblock \href {https://doi.org/10.1126/science.aaw2694} {The crisis of
  democracy and the science of deliberation}.
\newblock \emph{Science}, 363(6432):1144--1146.

\bibitem[{Durmus et~al.(2024)Durmus, Nguyen, Liao, Schiefer, Askell, Bakhtin,
  Chen, Hatfield-Dodds, Hernandez, Joseph, Lovitt, McCandlish, Sikder, Tamkin,
  Thamkul, Kaplan, Clark, and Ganguli}]{durmus2023towards}
Esin Durmus, Karina Nguyen, Thomas~I. Liao, Nicholas Schiefer, Amanda Askell,
  Anton Bakhtin, Carol Chen, Zac Hatfield-Dodds, Danny Hernandez, Nicholas
  Joseph, Liane Lovitt, Sam McCandlish, Orowa Sikder, Alex Tamkin, Janel
  Thamkul, Jared Kaplan, Jack Clark, and Deep Ganguli. 2024.
\newblock \href {http://arxiv.org/abs/2306.16388} {Towards measuring the
  representation of subjective global opinions in language models}.

\bibitem[{Ein-Dor et~al.(2020)Ein-Dor, Shnarch, Dankin, Halfon, Sznajder, Gera,
  Alzate, Gleize, Choshen, Hou, Bilu, Aharonov, and Slonim}]{ein2020corpus}
Liat Ein-Dor, Eyal Shnarch, Lena Dankin, Alon Halfon, Benjamin Sznajder, Ariel
  Gera, Carlos Alzate, Martin Gleize, Leshem Choshen, Yufang Hou, Yonatan Bilu,
  Ranit Aharonov, and Noam Slonim. 2020.
\newblock \href {https://doi.org/10.1609/aaai.v34i05.6270} {Corpus wide
  argument mining—a working solution}.
\newblock \emph{Proceedings of the AAAI Conference on Artificial Intelligence},
  34(05):7683--7691.

\bibitem[{Falk et~al.(2021)Falk, Jundi, Vecchi, and
  Lapesa}]{falk2021predicting}
Neele Falk, Iman Jundi, Eva~Maria Vecchi, and Gabriella Lapesa. 2021.
\newblock \href {https://doi.org/10.18653/v1/2021.argmining-1.13} {Predicting
  moderation of deliberative arguments: Is argument quality the key?}
\newblock In \emph{Proceedings of the 8th Workshop on Argument Mining}, pages
  133--141, Punta Cana, Dominican Republic. Association for Computational
  Linguistics.

\bibitem[{Feng et~al.(2023)Feng, Park, Liu, and Tsvetkov}]{feng2023pretraining}
Shangbin Feng, Chan~Young Park, Yuhan Liu, and Yulia Tsvetkov. 2023.
\newblock \href {https://doi.org/10.18653/v1/2023.acl-long.656} {From
  pretraining data to language models to downstream tasks: Tracking the trails
  of political biases leading to unfair {NLP} models}.
\newblock In \emph{Proceedings of the 61st Annual Meeting of the Association
  for Computational Linguistics (Volume 1: Long Papers)}, pages 11737--11762,
  Toronto, Canada. Association for Computational Linguistics.

\bibitem[{Fluri et~al.(2023)Fluri, Paleka, and
  Tram{\`e}r}]{fluri2023evaluating}
Lukas Fluri, Daniel Paleka, and Florian Tram{\`e}r. 2023.
\newblock \href {https://openreview.net/forum?id=L2ZIcu5fxS} {Evaluating
  superhuman models with consistency checks}.
\newblock In \emph{Socially Responsible Language Modelling Research}.

\bibitem[{Friess and Eilders(2015)}]{friess2015systematic}
Dennis Friess and Christiane Eilders. 2015.
\newblock \href {https://doi.org/10.1002/poi3.95} {A systematic review of
  online deliberation research}.
\newblock \emph{Policy \& Internet}, 7(3):319--339.

\bibitem[{Geirhos et~al.(2020)Geirhos, Jacobsen, Michaelis, Zemel, Brendel,
  Bethge, and Wichmann}]{geirhos2020shortcut}
Robert Geirhos, J{\"o}rn-Henrik Jacobsen, Claudio Michaelis, Richard Zemel,
  Wieland Brendel, Matthias Bethge, and Felix~A Wichmann. 2020.
\newblock \href {https://doi.org/10.1038/s42256-020-00257-z} {Shortcut learning
  in deep neural networks}.
\newblock \emph{Nature Machine Intelligence}, 2(11):665--673.

\bibitem[{Graham et~al.(2013)Graham, Haidt, Koleva, Motyl, Iyer, Wojcik, and
  Ditto}]{graham2013moral}
Jesse Graham, Jonathan Haidt, Sena Koleva, Matt Motyl, Ravi Iyer, Sean~P.
  Wojcik, and Peter~H. Ditto. 2013.
\newblock \href
  {https://doi.org/https://doi.org/10.1016/B978-0-12-407236-7.00002-4} {Chapter
  two - moral foundations theory: The pragmatic validity of moral pluralism}.
\newblock In Patricia Devine and Ashby Plant, editors, \emph{Advances in
  Experimental Social Psychology}, volume~47 of \emph{Advances in Experimental
  Social Psychology}, pages 55--130. Academic Press.

\bibitem[{Groarke(2024)}]{sep-logic-informal}
Leo Groarke. 2024.
\newblock \href
  {https://plato.stanford.edu/archives/spr2024/entries/logic-informal/}
  {{Informal Logic}}.
\newblock In Edward~N. Zalta and Uri Nodelman, editors, \emph{The {Stanford}
  Encyclopedia of Philosophy}, {S}pring 2024 edition. Metaphysics Research Lab,
  Stanford University.

\bibitem[{Habernal et~al.(2018)Habernal, Wachsmuth, Gurevych, and
  Stein}]{habernal2018argument}
Ivan Habernal, Henning Wachsmuth, Iryna Gurevych, and Benno Stein. 2018.
\newblock \href {https://doi.org/10.18653/v1/N18-1175} {The argument reasoning
  comprehension task: Identification and reconstruction of implicit warrants}.
\newblock In \emph{Proceedings of the 2018 Conference of the North {A}merican
  Chapter of the Association for Computational Linguistics: Human Language
  Technologies, Volume 1 (Long Papers)}, pages 1930--1940, New Orleans,
  Louisiana. Association for Computational Linguistics.

\bibitem[{Hartmann et~al.(2023)Hartmann, Schwenzow, and
  Witte}]{hartmann2023political}
Jochen Hartmann, Jasper Schwenzow, and Maximilian Witte. 2023.
\newblock \href {http://arxiv.org/abs/2301.01768} {The political ideology of
  conversational ai: Converging evidence on chatgpt's pro-environmental,
  left-libertarian orientation}.

\bibitem[{Heer(2019)}]{heer2019agency}
Jeffrey Heer. 2019.
\newblock \href {https://doi.org/10.1073/pnas.1807184115} {Agency plus
  automation: Designing artificial intelligence into interactive systems}.
\newblock \emph{Proceedings of the National Academy of Sciences},
  116(6):1844--1850.

\bibitem[{Hendrycks et~al.(2021)Hendrycks, Burns, Basart, Critch, Li, Song, and
  Steinhardt}]{hendrycks2020aligning}
Dan Hendrycks, Collin Burns, Steven Basart, Andrew Critch, Jerry Li, Dawn Song,
  and Jacob Steinhardt. 2021.
\newblock \href {https://openreview.net/forum?id=dNy_RKzJacY} {Aligning {AI}
  with shared human values}.
\newblock In \emph{International Conference on Learning Representations}.

\bibitem[{Hocevar et~al.(2014)Hocevar, Flanagin, and
  Metzger}]{hocevar2014social}
Kristin~Page Hocevar, Andrew~J. Flanagin, and Miriam~J. Metzger. 2014.
\newblock \href {https://doi.org/https://doi.org/10.1016/j.chb.2014.07.020}
  {Social media self-efficacy and information evaluation online}.
\newblock \emph{Computers in Human Behavior}, 39:254--262.

\bibitem[{Hoover et~al.(2020)Hoover, Portillo-Wightman, Yeh, Havaldar, Davani,
  Lin, Kennedy, Atari, Kamel, Mendlen, Moreno, Park, Chang, Chin, Leong, Leung,
  Mirinjian, and Dehghani}]{hoover2020moral}
Joe Hoover, Gwenyth Portillo-Wightman, Leigh Yeh, Shreya Havaldar,
  Aida~Mostafazadeh Davani, Ying Lin, Brendan Kennedy, Mohammad Atari, Zahra
  Kamel, Madelyn Mendlen, Gabriela Moreno, Christina Park, Tingyee~E. Chang,
  Jenna Chin, Christian Leong, Jun~Yen Leung, Arineh Mirinjian, and Morteza
  Dehghani. 2020.
\newblock \href {https://doi.org/10.1177/1948550619876629} {Moral foundations
  twitter corpus: A collection of 35k tweets annotated for moral sentiment}.
\newblock \emph{Social Psychological and Personality Science},
  11(8):1057--1071.

\bibitem[{Huang et~al.(2023)Huang, Yu, Ma, Zhong, Feng, Wang, Chen, Peng, Feng,
  Qin, and Liu}]{huang2023survey}
Lei Huang, Weijiang Yu, Weitao Ma, Weihong Zhong, Zhangyin Feng, Haotian Wang,
  Qianglong Chen, Weihua Peng, Xiaocheng Feng, Bing Qin, and Ting Liu. 2023.
\newblock \href {http://arxiv.org/abs/2311.05232} {A survey on hallucination in
  large language models: Principles, taxonomy, challenges, and open questions}.

\bibitem[{Iandoli et~al.(2014)Iandoli, Quinto, {De Liddo}, and {Buckingham
  Shum}}]{iandoli2014socially}
Luca Iandoli, Ivana Quinto, Anna {De Liddo}, and Simon {Buckingham Shum}. 2014.
\newblock \href {https://doi.org/https://doi.org/10.1016/j.ijhcs.2013.08.006}
  {Socially augmented argumentation tools: Rationale, design and evaluation of
  a debate dashboard}.
\newblock \emph{International Journal of Human-Computer Studies},
  72(3):298--319.

\bibitem[{Jelodar et~al.(2020)Jelodar, Wang, Orji, and Huang}]{jelodar2020deep}
Hamed Jelodar, Yongli Wang, Rita Orji, and Shucheng Huang. 2020.
\newblock \href {https://doi.org/10.1109/JBHI.2020.3001216} {Deep sentiment
  classification and topic discovery on novel coronavirus or {COVID-19} online
  discussions: {NLP} using {LSTM} recurrent neural network approach}.
\newblock \emph{IEEE Journal of Biomedical and Health Informatics},
  24(10):2733--2742.

\bibitem[{Jiang et~al.(2023)Jiang, Sablayrolles, Mensch, Bamford, Chaplot,
  de~las Casas, Bressand, Lengyel, Lample, Saulnier, Lavaud, Lachaux, Stock,
  Scao, Lavril, Wang, Lacroix, and Sayed}]{jiang2023mistral}
Albert~Q. Jiang, Alexandre Sablayrolles, Arthur Mensch, Chris Bamford,
  Devendra~Singh Chaplot, Diego de~las Casas, Florian Bressand, Gianna Lengyel,
  Guillaume Lample, Lucile Saulnier, Lélio~Renard Lavaud, Marie-Anne Lachaux,
  Pierre Stock, Teven~Le Scao, Thibaut Lavril, Thomas Wang, Timothée Lacroix,
  and William~El Sayed. 2023.
\newblock \href {http://arxiv.org/abs/2310.06825} {Mistral 7b}.

\bibitem[{Jo and Oh(2011)}]{jo2011aspect}
Yohan Jo and Alice~H. Oh. 2011.
\newblock \href {https://doi.org/10.1145/1935826.1935932} {Aspect and sentiment
  unification model for online review analysis}.
\newblock In \emph{Proceedings of the Fourth ACM International Conference on
  Web Search and Data Mining}, WSDM '11, page 815–824, New York, NY, USA.
  Association for Computing Machinery.

\bibitem[{Joshi et~al.(2020)Joshi, Santy, Budhiraja, Bali, and
  Choudhury}]{joshi2020state}
Pratik Joshi, Sebastin Santy, Amar Budhiraja, Kalika Bali, and Monojit
  Choudhury. 2020.
\newblock \href {https://doi.org/10.18653/v1/2020.acl-main.560} {The state and
  fate of linguistic diversity and inclusion in the {NLP} world}.
\newblock In \emph{Proceedings of the 58th Annual Meeting of the Association
  for Computational Linguistics}, pages 6282--6293, Online. Association for
  Computational Linguistics.

\bibitem[{Kiela et~al.(2021)Kiela, Bartolo, Nie, Kaushik, Geiger, Wu, Vidgen,
  Prasad, Singh, Ringshia, Ma, Thrush, Riedel, Waseem, Stenetorp, Jia, Bansal,
  Potts, and Williams}]{kiela-etal-2021-dynabench}
Douwe Kiela, Max Bartolo, Yixin Nie, Divyansh Kaushik, Atticus Geiger,
  Zhengxuan Wu, Bertie Vidgen, Grusha Prasad, Amanpreet Singh, Pratik Ringshia,
  Zhiyi Ma, Tristan Thrush, Sebastian Riedel, Zeerak Waseem, Pontus Stenetorp,
  Robin Jia, Mohit Bansal, Christopher Potts, and Adina Williams. 2021.
\newblock \href {https://doi.org/10.18653/v1/2021.naacl-main.324} {Dynabench:
  Rethinking benchmarking in {NLP}}.
\newblock In \emph{Proceedings of the 2021 Conference of the North American
  Chapter of the Association for Computational Linguistics: Human Language
  Technologies}, pages 4110--4124, Online. Association for Computational
  Linguistics.

\bibitem[{Kiesel et~al.(2022)Kiesel, Alshomary, Handke, Cai, Wachsmuth, and
  Stein}]{kiesel2022identifying}
Johannes Kiesel, Milad Alshomary, Nicolas Handke, Xiaoni Cai, Henning
  Wachsmuth, and Benno Stein. 2022.
\newblock \href {https://doi.org/10.18653/v1/2022.acl-long.306} {Identifying
  the human values behind arguments}.
\newblock In \emph{Proceedings of the 60th Annual Meeting of the Association
  for Computational Linguistics (Volume 1: Long Papers)}, pages 4459--4471,
  Dublin, Ireland. Association for Computational Linguistics.

\bibitem[{Kiesel et~al.(2023)Kiesel, Alshomary, Mirzakhmedova, Heinrich,
  Handke, Wachsmuth, and Stein}]{kiesel2023semeval}
Johannes Kiesel, Milad Alshomary, Nailia Mirzakhmedova, Maximilian Heinrich,
  Nicolas Handke, Henning Wachsmuth, and Benno Stein. 2023.
\newblock \href {https://doi.org/10.18653/v1/2023.semeval-1.313}
  {{S}em{E}val-2023 task 4: {V}alue{E}val: Identification of human values
  behind arguments}.
\newblock In \emph{Proceedings of the 17th International Workshop on Semantic
  Evaluation (SemEval-2023)}, pages 2287--2303, Toronto, Canada. Association
  for Computational Linguistics.

\bibitem[{Klein(2012)}]{klein2012enabling}
Mark Klein. 2012.
\newblock \href {https://doi.org/10.1007/s10606-012-9156-4} {Enabling
  large-scale deliberation using attention-mediation metrics}.
\newblock \emph{Computer Supported Cooperative Work (CSCW)}, 21:449--473.

\bibitem[{Knobel and Lankshear(2008)}]{knobel2008digital}
Michele Knobel and Colin Lankshear. 2008.
\newblock \href
  {https://pages.ucsd.edu/~bgoldfarb/comt109w10/reading/Lankshear-Knobel_et_al-DigitalLiteracies.pdf#page=251}
  {Digital literacy and participation in online social networking spaces}.
\newblock \emph{Digital literacies: Concepts, policies and practices},
  11:249--278.

\bibitem[{Kong et~al.(2022)Kong, Booth, Bailo, Johns, and
  Rizoiu}]{kong2022slipping}
Quyu Kong, Emily Booth, Francesco Bailo, Amelia Johns, and Marian-Andrei
  Rizoiu. 2022.
\newblock \href {https://doi.org/10.1609/icwsm.v16i1.19312} {Slipping to the
  extreme: A mixed method to explain how extreme opinions infiltrate online
  discussions}.
\newblock \emph{Proceedings of the International AAAI Conference on Web and
  Social Media}, 16(1):524--535.

\bibitem[{K\"{u}\c{c}\"{u}k and Can(2020)}]{kuccuk2020stance}
Dilek K\"{u}\c{c}\"{u}k and Fazli Can. 2020.
\newblock \href {https://doi.org/10.1145/3369026} {Stance detection: A survey}.
\newblock \emph{ACM Comput. Surv.}, 53(1).

\bibitem[{Laban et~al.(2022)Laban, Wu, Murakhovs{'}ka, Chen, and
  Xiong}]{laban2022discord}
Philippe Laban, Chien-Sheng Wu, Lidiya Murakhovs{'}ka, Xiang Chen, and Caiming
  Xiong. 2022.
\newblock \href {https://doi.org/10.18653/v1/2022.findings-emnlp.380} {Discord
  questions: A computational approach to diversity analysis in news coverage}.
\newblock In \emph{Findings of the Association for Computational Linguistics:
  EMNLP 2022}, pages 5180--5194, Abu Dhabi, United Arab Emirates. Association
  for Computational Linguistics.

\bibitem[{Lahoti et~al.(2023)Lahoti, Blumm, Ma, Kotikalapudi, Potluri, Tan,
  Srinivasan, Packer, Beirami, Beutel, and Chen}]{lahoti2023improving}
Preethi Lahoti, Nicholas Blumm, Xiao Ma, Raghavendra Kotikalapudi, Sahitya
  Potluri, Qijun Tan, Hansa Srinivasan, Ben Packer, Ahmad Beirami, Alex Beutel,
  and Jilin Chen. 2023.
\newblock \href {https://doi.org/10.18653/v1/2023.emnlp-main.643} {Improving
  diversity of demographic representation in large language models via
  collective-critiques and self-voting}.
\newblock In \emph{Proceedings of the 2023 Conference on Empirical Methods in
  Natural Language Processing}, pages 10383--10405, Singapore. Association for
  Computational Linguistics.

\bibitem[{Lapesa et~al.(2023)Lapesa, Vecchi, Villata, and
  Wachsmuth}]{lapesa2023mining}
Gabriella Lapesa, Eva~Maria Vecchi, Serena Villata, and Henning Wachsmuth.
  2023.
\newblock \href {https://doi.org/10.18653/v1/2023.eacl-tutorials.1} {Mining,
  assessing, and improving arguments in {NLP} and the social sciences}.
\newblock In \emph{Proceedings of the 17th Conference of the European Chapter
  of the Association for Computational Linguistics: Tutorial Abstracts}, pages
  1--6, Dubrovnik, Croatia. Association for Computational Linguistics.

\bibitem[{Lawrence and Reed(2020)}]{lawrence2020argument}
John Lawrence and Chris Reed. 2020.
\newblock \href {https://doi.org/10.1162/coli_a_00364} {{Argument Mining: A
  Survey}}.
\newblock \emph{Computational Linguistics}, 45(4):765--818.

\bibitem[{Lazer et~al.(2009)Lazer, Pentland, Adamic, Aral, Barabási, Brewer,
  Christakis, Contractor, Fowler, Gutmann, Jebara, King, Macy, Roy, and
  Alstyne}]{lazer2009computational}
David Lazer, Alex Pentland, Lada Adamic, Sinan Aral, Albert-László Barabási,
  Devon Brewer, Nicholas Christakis, Noshir Contractor, James Fowler, Myron
  Gutmann, Tony Jebara, Gary King, Michael Macy, Deb Roy, and Marshall~Van
  Alstyne. 2009.
\newblock \href {https://doi.org/10.1126/science.1167742} {Computational social
  science}.
\newblock \emph{Science}, 323(5915):721--723.

\bibitem[{Lertvittayakumjorn and
  Toni(2021)}]{lertvittayakumjorn2021explanation}
Piyawat Lertvittayakumjorn and Francesca Toni. 2021.
\newblock \href {https://doi.org/10.1162/tacl_a_00440} {{Explanation-Based
  Human Debugging of {NLP} Models: A Survey}}.
\newblock \emph{Transactions of the Association for Computational Linguistics},
  9:1508--1528.

\bibitem[{Liang et~al.(2021)Liang, Wu, Morency, and
  Salakhutdinov}]{liang2021towards}
Paul~Pu Liang, Chiyu Wu, Louis-Philippe Morency, and Ruslan Salakhutdinov.
  2021.
\newblock \href {https://proceedings.mlr.press/v139/liang21a.html} {Towards
  understanding and mitigating social biases in language models}.
\newblock In \emph{Proceedings of the 38th International Conference on Machine
  Learning}, volume 139 of \emph{Proceedings of Machine Learning Research},
  pages 6565--6576. PMLR.

\bibitem[{Lin et~al.(2009)Lin, Hsieh, and Chuang}]{lin2009discovering}
Fu-Ren Lin, Lu-Shih Hsieh, and Fu-Tai Chuang. 2009.
\newblock \href {https://doi.org/https://doi.org/10.1016/j.compedu.2008.10.005}
  {Discovering genres of online discussion threads via text mining}.
\newblock \emph{Computers \& Education}, 52(2):481--495.

\bibitem[{Liscio et~al.(2023)Liscio, Lera-Leri, Bistaffa, Dobbe, Jonker,
  Lopez-Sanchez, Rodriguez-Aguilar, and Murukannaiah}]{liscio2023value}
Enrico Liscio, Roger Lera-Leri, Filippo Bistaffa, Roel~I.J. Dobbe, Catholijn~M.
  Jonker, Maite Lopez-Sanchez, Juan~A. Rodriguez-Aguilar, and Pradeep~K.
  Murukannaiah. 2023.
\newblock \href {https://dl.acm.org/doi/abs/10.5555/3545946.3598838} {Value
  inference in sociotechnical systems}.
\newblock In \emph{Proceedings of the 2023 International Conference on
  Autonomous Agents and Multiagent Systems}, AAMAS '23, page 1774–1780,
  Richland, SC. International Foundation for Autonomous Agents and Multiagent
  Systems.

\bibitem[{Liscio et~al.(2024)Liscio, Siebert, Jonker, and
  Murukannaiah}]{liscio2024value}
Enrico Liscio, Luciano~C. Siebert, Catholijn~M. Jonker, and Pradeep~K.
  Murukannaiah. 2024.
\newblock \href {http://arxiv.org/abs/2402.16751} {Value preferences estimation
  and disambiguation in hybrid participatory systems}.

\bibitem[{Liu(2012)}]{liu2012sentiment}
Bing Liu. 2012.
\newblock \href {https://doi.org/10.1007/978-3-031-02145-9} {\emph{Sentiment
  Analysis and Opinion Mining}}.
\newblock Springer International Publishing.

\bibitem[{L{\"o}rcher and Taddicken(2017)}]{lorcher2017discussing}
Ines L{\"o}rcher and Monika Taddicken. 2017.
\newblock \href
  {https://pure.mpg.de/pubman/faces/ViewItemOverviewPage.jsp?itemId=item_2473044}
  {Discussing climate change online. topics and perceptions in online climate
  change communication in different online public arenas}.
\newblock \emph{Journal of Science Communication}, 16(2):A03.

\bibitem[{Lorenz et~al.(2011)Lorenz, Rauhut, Schweitzer, and
  Helbing}]{lorenz2011social}
Jan Lorenz, Heiko Rauhut, Frank Schweitzer, and Dirk Helbing. 2011.
\newblock \href {https://doi.org/10.1073/pnas.1008636108} {How social influence
  can undermine the wisdom of crowd effect}.
\newblock \emph{Proceedings of the National Academy of Sciences},
  108(22):9020--9025.

\bibitem[{Lynn et~al.(2017)Lynn, Son, Kulkarni, Balasubramanian, and
  Schwartz}]{lynn2017human}
Veronica Lynn, Youngseo Son, Vivek Kulkarni, Niranjan Balasubramanian, and
  H.~Andrew Schwartz. 2017.
\newblock \href {https://doi.org/10.18653/v1/D17-1119} {Human centered {NLP}
  with user-factor adaptation}.
\newblock In \emph{Proceedings of the 2017 Conference on Empirical Methods in
  Natural Language Processing}, pages 1146--1155, Copenhagen, Denmark.
  Association for Computational Linguistics.

\bibitem[{Madras et~al.(2018)Madras, Pitassi, and Zemel}]{madras2018predict}
David Madras, Toni Pitassi, and Richard Zemel. 2018.
\newblock \href
  {https://proceedings.neurips.cc/paper_files/paper/2018/file/09d37c08f7b129e96277388757530c72-Paper.pdf}
  {Predict responsibly: Improving fairness and accuracy by learning to defer}.
\newblock In \emph{Advances in Neural Information Processing Systems},
  volume~31. Curran Associates, Inc.

\bibitem[{Min et~al.(2023)Min, Ross, Sulem, Veyseh, Nguyen, Sainz, Agirre,
  Heintz, and Roth}]{min2023recent}
Bonan Min, Hayley Ross, Elior Sulem, Amir Pouran~Ben Veyseh, Thien~Huu Nguyen,
  Oscar Sainz, Eneko Agirre, Ilana Heintz, and Dan Roth. 2023.
\newblock \href {https://doi.org/10.1145/3605943} {Recent advances in natural
  language processing via large pre-trained language models: A survey}.
\newblock \emph{ACM Comput. Surv.}, 56(2).

\bibitem[{Mooijman et~al.(2018)Mooijman, Hoover, Lin, Ji, and
  Dehghani}]{mooijman2018moralization}
Marlon Mooijman, Joe Hoover, Ying Lin, Heng Ji, and Morteza Dehghani. 2018.
\newblock \href {https://doi.org/10.1038/s41562-018-0353-0} {Moralization in
  social networks and the emergence of violence during protests}.
\newblock \emph{Nature human behaviour}, 2(6):389--396.

\bibitem[{Morante et~al.(2020)Morante, van Son, Maks, and
  Vossen}]{morante2020annotating}
Roser Morante, Chantal van Son, Isa Maks, and Piek Vossen. 2020.
\newblock \href {https://aclanthology.org/2020.lrec-1.611} {Annotating
  perspectives on vaccination}.
\newblock In \emph{Proceedings of the Twelfth Language Resources and Evaluation
  Conference}, pages 4964--4973, Marseille, France. European Language Resources
  Association.

\bibitem[{Mouter et~al.(2021)Mouter, Hernandez, and Itten}]{mouter2021public}
Niek Mouter, Jose~Ignacio Hernandez, and Anatol~Valerian Itten. 2021.
\newblock \href {https://doi.org/10.1371/journal.pone.0250614} {Public
  participation in crisis policymaking. how 30,000 dutch citizens advised their
  government on relaxing covid-19 lockdown measures}.
\newblock \emph{PLOS ONE}, 16(5):1--42.

\bibitem[{Mustafaraj et~al.(2011)Mustafaraj, Finn, Whitlock, and
  Metaxas}]{mustafaraj2011vocal}
Eni Mustafaraj, Samantha Finn, Carolyn Whitlock, and Panagiotis~T. Metaxas.
  2011.
\newblock \href {https://doi.org/10.1109/PASSAT/SocialCom.2011.188} {Vocal
  minority versus silent majority: Discovering the opionions of the long tail}.
\newblock In \emph{2011 IEEE Third International Conference on Privacy,
  Security, Risk and Trust and 2011 IEEE Third International Conference on
  Social Computing}, pages 103--110.

\bibitem[{Neubaum and Krämer(2017)}]{neubaum2017monitoring}
German Neubaum and Nicole~C. Krämer. 2017.
\newblock \href {https://doi.org/10.1080/15213269.2016.1211539} {Monitoring the
  opinion of the crowd: Psychological mechanisms underlying public opinion
  perceptions on social media}.
\newblock \emph{Media Psychology}, 20(3):502--531.

\bibitem[{Nie et~al.(2020)Nie, Zhou, and Bansal}]{nie2020learn}
Yixin Nie, Xiang Zhou, and Mohit Bansal. 2020.
\newblock \href {https://doi.org/10.18653/v1/2020.emnlp-main.734} {What can we
  learn from collective human opinions on natural language inference data?}
\newblock In \emph{Proceedings of the 2020 Conference on Empirical Methods in
  Natural Language Processing (EMNLP)}, pages 9131--9143, Online. Association
  for Computational Linguistics.

\bibitem[{Ouyang et~al.(2023)Ouyang, Wang, Liu, Zhong, Jiao, Iter, Pryzant,
  Zhu, Ji, and Han}]{ouyang2023shifted}
Siru Ouyang, Shuohang Wang, Yang Liu, Ming Zhong, Yizhu Jiao, Dan Iter, Reid
  Pryzant, Chenguang Zhu, Heng Ji, and Jiawei Han. 2023.
\newblock \href {https://doi.org/10.18653/v1/2023.emnlp-main.146} {The shifted
  and the overlooked: A task-oriented investigation of user-{GPT}
  interactions}.
\newblock In \emph{Proceedings of the 2023 Conference on Empirical Methods in
  Natural Language Processing}, pages 2375--2393, Singapore. Association for
  Computational Linguistics.

\bibitem[{Pacheco et~al.(2023)Pacheco, Islam, Ungar, Yin, and
  Goldwasser}]{pacheco2023interactive}
Maria~Leonor Pacheco, Tunazzina Islam, Lyle Ungar, Ming Yin, and Dan
  Goldwasser. 2023.
\newblock \href {https://doi.org/10.18653/v1/2023.findings-acl.313}
  {Interactive concept learning for uncovering latent themes in large text
  collections}.
\newblock In \emph{Findings of the Association for Computational Linguistics:
  ACL 2023}, pages 5059--5080, Toronto, Canada. Association for Computational
  Linguistics.

\bibitem[{Plank(2022)}]{plank2022problem}
Barbara Plank. 2022.
\newblock \href {https://doi.org/10.18653/v1/2022.emnlp-main.731} {The
  {``}problem{''} of human label variation: On ground truth in data, modeling
  and evaluation}.
\newblock In \emph{Proceedings of the 2022 Conference on Empirical Methods in
  Natural Language Processing}, pages 10671--10682, Abu Dhabi, United Arab
  Emirates. Association for Computational Linguistics.

\bibitem[{Ponizovskiy et~al.(2020)Ponizovskiy, Ardag, Grigoryan, Boyd,
  Dobewall, and Holtz}]{ponizovskiy2020development}
Vladimir Ponizovskiy, Murat Ardag, Lusine Grigoryan, Ryan Boyd, Henrik
  Dobewall, and Peter Holtz. 2020.
\newblock \href {https://doi.org/10.1002/per.2294} {Development and validation
  of the personal values dictionary: A theory–driven tool for investigating
  references to basic human values in text}.
\newblock \emph{European Journal of Personality}, 34(5):885--902.

\bibitem[{Pougu{\'e}-Biyong et~al.(2021)Pougu{\'e}-Biyong, Semenova, Matton,
  Han, Kim, Lambiotte, and Farmer}]{pougue2021debagreement}
John Pougu{\'e}-Biyong, Valentina Semenova, Alexandre Matton, Rachel Han, Aerin
  Kim, Renaud Lambiotte, and Doyne Farmer. 2021.
\newblock \href {https://openreview.net/forum?id=udVUN__gFO} {{DEBAGREEMENT}: A
  comment-reply dataset for (dis)agreement detection in online debates}.
\newblock In \emph{Thirty-fifth Conference on Neural Information Processing
  Systems Datasets and Benchmarks Track (Round 2)}.

\bibitem[{Reimers and Gurevych(2019)}]{reimers2019sentence}
Nils Reimers and Iryna Gurevych. 2019.
\newblock \href {https://doi.org/10.18653/v1/D19-1410} {Sentence-{BERT}:
  Sentence embeddings using {S}iamese {BERT}-networks}.
\newblock In \emph{Proceedings of the 2019 Conference on Empirical Methods in
  Natural Language Processing and the 9th International Joint Conference on
  Natural Language Processing (EMNLP-IJCNLP)}, pages 3982--3992, Hong Kong,
  China. Association for Computational Linguistics.

\bibitem[{Renting et~al.(2022)Renting, Hoos, and Jonker}]{renting2022automated}
Bram~M. Renting, Holger~H. Hoos, and Catholijn~M. Jonker. 2022.
\newblock \href {https://dl.acm.org/doi/abs/10.5555/3535850.3535973} {Automated
  configuration and usage of strategy portfolios for mixed-motive bargaining}.
\newblock In \emph{Proceedings of the 21st International Conference on
  Autonomous Agents and Multiagent Systems}, AAMAS '22, page 1101–1109,
  Richland, SC. International Foundation for Autonomous Agents and Multiagent
  Systems.

\bibitem[{Rokeach(1967)}]{rokeach1967rokeach}
Milton Rokeach. 1967.
\newblock \href {https://doi.org/10.1037/t01381-000} {Rokeach value survey}.
\newblock \emph{The nature of human values.}

\bibitem[{Roy et~al.(2021)Roy, Pacheco, and Goldwasser}]{roy2021identifying}
Shamik Roy, Maria~Leonor Pacheco, and Dan Goldwasser. 2021.
\newblock \href {https://doi.org/10.18653/v1/2021.emnlp-main.783} {Identifying
  morality frames in political tweets using relational learning}.
\newblock In \emph{Proceedings of the 2021 Conference on Empirical Methods in
  Natural Language Processing}, pages 9939--9958, Online and Punta Cana,
  Dominican Republic. Association for Computational Linguistics.

\bibitem[{Santamar\'{i}a et~al.(2022)Santamar\'{i}a, Vossen, and
  Baier}]{santamaria2022evaluating}
Selene~B\'{a}ez Santamar\'{i}a, Piek Vossen, and Thomas Baier. 2022.
\newblock \href {https://aclanthology.org/2022.ccgpk-1.3} {Evaluating agent
  interactions through episodic knowledge graphs}.
\newblock In \emph{Proceedings of the 1st Workshop on Customized Chat Grounding
  Persona and Knowledge}, pages 15--28, Gyeongju, Republic of Korea.
  Association for Computational Linguistics.

\bibitem[{Santurkar et~al.(2023)Santurkar, Durmus, Ladhak, Lee, Liang, and
  Hashimoto}]{santurkar2023whose}
Shibani Santurkar, Esin Durmus, Faisal Ladhak, Cinoo Lee, Percy Liang, and
  Tatsunori Hashimoto. 2023.
\newblock \href {https://proceedings.mlr.press/v202/santurkar23a.html} {Whose
  opinions do language models reflect?}
\newblock In \emph{Proceedings of the 40th International Conference on Machine
  Learning}, volume 202 of \emph{Proceedings of Machine Learning Research},
  pages 29971--30004. PMLR.

\bibitem[{Santy et~al.(2023)Santy, Liang, Le~Bras, Reinecke, and
  Sap}]{santy2023nlpositionality}
Sebastin Santy, Jenny Liang, Ronan Le~Bras, Katharina Reinecke, and Maarten
  Sap. 2023.
\newblock \href {https://doi.org/10.18653/v1/2023.acl-long.505}
  {{NLP}ositionality: Characterizing design biases of datasets and models}.
\newblock In \emph{Proceedings of the 61st Annual Meeting of the Association
  for Computational Linguistics (Volume 1: Long Papers)}, pages 9080--9102,
  Toronto, Canada. Association for Computational Linguistics.

\bibitem[{Saxena and Reddy(2022)}]{saxena2022users}
Akrati Saxena and Harita Reddy. 2022.
\newblock \href {https://doi.org/10.1007/s42001-021-00125-9} {Users roles
  identification on online crowdsourced {Q\&} platforms and encyclopedias: a
  survey}.
\newblock \emph{Journal of Computational Social Science}, 5(1):285--317.

\bibitem[{Schaeffer et~al.(2023)Schaeffer, Miranda, and
  Koyejo}]{schaeffer2024emergent}
Rylan Schaeffer, Brando Miranda, and Sanmi Koyejo. 2023.
\newblock \href
  {https://proceedings.neurips.cc/paper_files/paper/2023/file/adc98a266f45005c403b8311ca7e8bd7-Paper-Conference.pdf}
  {Are emergent abilities of large language models a mirage?}
\newblock In \emph{Advances in Neural Information Processing Systems},
  volume~36, pages 55565--55581. Curran Associates, Inc.

\bibitem[{Schwartz(2012)}]{schwartz2012overview}
Shalom~H Schwartz. 2012.
\newblock \href {https://doi.org/10.9707/2307-0919.1116} {An overview of the
  schwartz theory of basic values}.
\newblock \emph{Online readings in Psychology and Culture}, 2(1):11.

\bibitem[{Shortall et~al.(2022)Shortall, Itten, van~der Meer, Murukannaiah, and
  Jonker}]{shortall2022reason}
Ruth Shortall, Anatol Itten, Michiel van~der Meer, Pradeep Murukannaiah, and
  Catholijn Jonker. 2022.
\newblock \href
  {https://www.frontiersin.org/articles/10.3389/fpos.2022.946589/full} {Reason
  against the machine? {F}uture directions for mass online deliberation}.
\newblock \emph{Frontiers in Political Science}.

\bibitem[{Skiera et~al.(2022)Skiera, Yan, Daxenberger, Dombois, and
  Gurevych}]{skiera2022using}
Bernd Skiera, Shunyao Yan, Johannes Daxenberger, Marcus Dombois, and Iryna
  Gurevych. 2022.
\newblock \href {https://doi.org/10.1177/10946705221110845} {Using
  information-seeking argument mining to improve service}.
\newblock \emph{Journal of Service Research}, 25(4):537--548.

\bibitem[{Smith(2009)}]{smith2009democratic}
G.~Smith. 2009.
\newblock \href {https://books.google.nl/books?id=gz8gAwAAQBAJ}
  {\emph{Democratic Innovations: Designing Institutions for Citizen
  Participation}}.
\newblock Theories of Institutional Design. Cambridge University Press.

\bibitem[{Song et~al.(2020)Song, Cho, and Benefield}]{song2020dynamics}
Hyunjin Song, Jaeho Cho, and Grace~A. Benefield. 2020.
\newblock \href {https://doi.org/10.1177/0093650218790144} {The dynamics of
  message selection in online political discussion forums: Self-segregation or
  diverse exposure?}
\newblock \emph{Communication Research}, 47(1):125--152.

\bibitem[{Steenbergen et~al.(2003)Steenbergen, B{\"a}chtiger, Sp{\"o}rndli, and
  Steiner}]{steenbergen2003measuring}
Marco~R Steenbergen, Andr{\'e} B{\"a}chtiger, Markus Sp{\"o}rndli, and J{\"u}rg
  Steiner. 2003.
\newblock \href {https://doi.org/10.1057/palgrave.cep.6110002} {Measuring
  political deliberation: A discourse quality index}.
\newblock \emph{Comparative European Politics}, 1:21--48.

\bibitem[{Sun et~al.(2017)Sun, Luo, and Chen}]{sun2017review}
Shiliang Sun, Chen Luo, and Junyu Chen. 2017.
\newblock \href {https://doi.org/https://doi.org/10.1016/j.inffus.2016.10.004}
  {A review of natural language processing techniques for opinion mining
  systems}.
\newblock \emph{Information Fusion}, 36:10--25.

\bibitem[{Surowiecki(2004)}]{surowiecki2004wisdom}
James Surowiecki. 2004.
\newblock \href {https://psycnet.apa.org/record/2004-20179-000} {\emph{The
  wisdom of crowds: Why the many are smarter than the few and how collective
  wisdom shapes business, economies, societies, and nations.}}
\newblock Doubleday.

\bibitem[{Syed et~al.(2023)Syed, Schwabe, Al-Khatib, and
  Potthast}]{syed2023indicative}
Shahbaz Syed, Dominik Schwabe, Khalid Al-Khatib, and Martin Potthast. 2023.
\newblock \href {https://doi.org/10.18653/v1/2023.emnlp-main.166} {Indicative
  summarization of long discussions}.
\newblock In \emph{Proceedings of the 2023 Conference on Empirical Methods in
  Natural Language Processing}, pages 2752--2788, Singapore. Association for
  Computational Linguistics.

\bibitem[{Team et~al.(2022)Team, Costa-jussà, Cross, Çelebi, Elbayad,
  Heafield, Heffernan, Kalbassi, Lam, Licht, Maillard, Sun, Wang, Wenzek,
  Youngblood, Akula, Barrault, Gonzalez, Hansanti, Hoffman, Jarrett, Sadagopan,
  Rowe, Spruit, Tran, Andrews, Ayan, Bhosale, Edunov, Fan, Gao, Goswami,
  Guzmán, Koehn, Mourachko, Ropers, Saleem, Schwenk, and Wang}]{costa2022no}
NLLB Team, Marta~R. Costa-jussà, James Cross, Onur Çelebi, Maha Elbayad,
  Kenneth Heafield, Kevin Heffernan, Elahe Kalbassi, Janice Lam, Daniel Licht,
  Jean Maillard, Anna Sun, Skyler Wang, Guillaume Wenzek, Al~Youngblood, Bapi
  Akula, Loic Barrault, Gabriel~Mejia Gonzalez, Prangthip Hansanti, John
  Hoffman, Semarley Jarrett, Kaushik~Ram Sadagopan, Dirk Rowe, Shannon Spruit,
  Chau Tran, Pierre Andrews, Necip~Fazil Ayan, Shruti Bhosale, Sergey Edunov,
  Angela Fan, Cynthia Gao, Vedanuj Goswami, Francisco Guzmán, Philipp Koehn,
  Alexandre Mourachko, Christophe Ropers, Safiyyah Saleem, Holger Schwenk, and
  Jeff Wang. 2022.
\newblock \href {http://arxiv.org/abs/2207.04672} {No language left behind:
  Scaling human-centered machine translation}.

\bibitem[{Tiddi et~al.(2023)Tiddi, de~Boer, Schlobach, and
  Meyer-Vitali}]{tiddi2023knowledge}
Ilaria Tiddi, Victor de~Boer, Stefan Schlobach, and André Meyer-Vitali. 2023.
\newblock \href {https://doi.org/10.1145/3587259.3627541} {Knowledge
  engineering for hybrid intelligence}.
\newblock In \emph{Proceedings of the 12th Knowledge Capture Conference 2023},
  K-CAP '23, page 75–82, Pensacola, FL, USA,. Association for Computing
  Machinery.

\bibitem[{Toma{\v{s}}ev et~al.(2020)Toma{\v{s}}ev, Cornebise, Hutter, Mohamed,
  Picciariello, Connelly, Belgrave, Ezer, Haert, Mugisha
  et~al.}]{tomavsev2020ai}
Nenad Toma{\v{s}}ev, Julien Cornebise, Frank Hutter, Shakir Mohamed, Angela
  Picciariello, Bec Connelly, Danielle~CM Belgrave, Daphne Ezer, Fanny Cachat
  van~der Haert, Frank Mugisha, et~al. 2020.
\newblock \href {https://doi.org/10.1038/s41467-020-15871-z} {{AI} for social
  good: unlocking the opportunity for positive impact}.
\newblock \emph{Nature Communications}, 11(1):2468.

\bibitem[{Toulmin(2003)}]{toulmin2003uses}
S.E. Toulmin. 2003.
\newblock \href {https://books.google.nl/books?id=8UYgegaB1S0C} {\emph{The Uses
  of Argument}}.
\newblock Cambridge University Press.

\bibitem[{Tr{\'e}nel(2009)}]{trenel2009facilitation}
Matthias Tr{\'e}nel. 2009.
\newblock \href
  {https://www.researchgate.net/profile/Todd-Davies-2/publication/37705170_Online_Deliberation_Design_Research_and_Practice/links/576b194d08ae5b9a62b3a7f8/Online-Deliberation-Design-Research-and-Practice.pdf#page=266}
  {Facilitation and inclusive deliberation}.
\newblock \emph{Online deliberation: Design, research, and practice}, pages
  253--257.

\bibitem[{Uma et~al.(2021)Uma, Fornaciari, Hovy, Paun, Plank, and
  Poesio}]{uma2021learning}
Alexandra~N Uma, Tommaso Fornaciari, Dirk Hovy, Silviu Paun, Barbara Plank, and
  Massimo Poesio. 2021.
\newblock \href {https://doi.org/10.1613/jair.1.12752} {Learning from
  disagreement: A survey}.
\newblock \emph{Journal of Artificial Intelligence Research}, 72:1385--1470.

\bibitem[{van~de Ven et~al.(2022)van~de Ven, Tuytelaars, and
  Tolias}]{van2022three}
Gido~M van~de Ven, Tinne Tuytelaars, and Andreas~S Tolias. 2022.
\newblock \href {https://doi.org/10.1038/s42256-022-00568-3} {Three types of
  incremental learning}.
\newblock \emph{Nature Machine Intelligence}, 4(12):1185--1197.

\bibitem[{van~der Meer et~al.(2024{\natexlab{a}})van~der Meer, Falk,
  Murukannaiah, and Liscio}]{van2024annotator}
Michiel van~der Meer, Neele Falk, Pradeep~K. Murukannaiah, and Enrico Liscio.
  2024{\natexlab{a}}.
\newblock \href {http://arxiv.org/abs/2404.15720} {Annotator-centric active
  learning for subjective {NLP} tasks}.

\bibitem[{van~der Meer et~al.(2022)van~der Meer, Liscio, Jonker, Plaat, Vossen,
  and Murukannaiah}]{vandermeer2022hyena}
Michiel van~der Meer, Enrico Liscio, Catholijn~M. Jonker, Aske Plaat, Piek
  Vossen, and Pradeep~K. Murukannaiah. 2022.
\newblock \href {https://liacs.leidenuniv.nl/~meermtvander/publications/hyena/}
  {{HyEnA: A Hybrid Method for Extracting Arguments from Opinions}}.
\newblock In \emph{Proceedings of the first International Conference on Hybrid
  Human-Artificial Intelligence (HHAI 2022)}, pages 1--15, Amsterdam, the
  Netherlands. IOS Press.

\bibitem[{van~der Meer et~al.(2024{\natexlab{b}})van~der Meer, Liscio, Jonker,
  Plaat, Vossen, and Murukannaiah}]{van2024hybrid}
Michiel van~der Meer, Enrico Liscio, Catholijn~M Jonker, Aske Plaat, Piek
  Vossen, and Pradeep~K Murukannaiah. 2024{\natexlab{b}}.
\newblock A hybrid intelligence method for argument mining.
\newblock \emph{Journal of AI Research (JAIR, to appear)}.

\bibitem[{van~der Meer et~al.(2023)van~der Meer, Vossen, Jonker, and
  Murukannaiah}]{vandermeer2023differences}
Michiel van~der Meer, Piek Vossen, Catholijn Jonker, and Pradeep Murukannaiah.
  2023.
\newblock \href {https://doi.org/10.18653/v1/2023.emnlp-main.992} {Do
  differences in values influence disagreements in online discussions?}
\newblock In \emph{Proceedings of the 2023 Conference on Empirical Methods in
  Natural Language Processing}, pages 15986--16008, Singapore. Association for
  Computational Linguistics.

\bibitem[{van~der Meer et~al.(2024{\natexlab{c}})van~der Meer, Vossen, Jonker,
  and Murukannaiah}]{vandermeer2024empirical}
Michiel van~der Meer, Piek Vossen, Catholijn Jonker, and Pradeep Murukannaiah.
  2024{\natexlab{c}}.
\newblock \href {https://aclanthology.org/2024.eacl-long.123} {An empirical
  analysis of diversity in argument summarization}.
\newblock In \emph{Proceedings of the 18th Conference of the European Chapter
  of the Association for Computational Linguistics (Volume 1: Long Papers)},
  pages 2028--2045, St. Julian{'}s, Malta. Association for Computational
  Linguistics.

\bibitem[{Van~Eemeren et~al.(2015)Van~Eemeren, van Eemeren, Jackson, and
  Jacobs}]{van2015argumentation}
Frans~H Van~Eemeren, Frans~H van Eemeren, Sally Jackson, and Scott Jacobs.
  2015.
\newblock Argumentation.
\newblock \emph{Reasonableness and effectiveness in argumentative discourse:
  Fifty contributions to the development of Pragma-dialectics}, pages 3--25.

\bibitem[{van Son et~al.(2016)van Son, Caselli, Fokkens, Maks, Morante, Aroyo,
  and Vossen}]{vanson2016grasp}
Chantal van Son, Tommaso Caselli, Antske Fokkens, Isa Maks, Roser Morante, Lora
  Aroyo, and Piek Vossen. 2016.
\newblock \href {https://aclanthology.org/L16-1187} {{GR}a{SP}: A multilayered
  annotation scheme for perspectives}.
\newblock In \emph{Proceedings of the Tenth International Conference on
  Language Resources and Evaluation ({LREC}'16)}, pages 1177--1184,
  Portoro{\v{z}}, Slovenia. European Language Resources Association (ELRA).

\bibitem[{Vecchi et~al.(2021)Vecchi, Falk, Jundi, and
  Lapesa}]{vecchi2021towards}
Eva~Maria Vecchi, Neele Falk, Iman Jundi, and Gabriella Lapesa. 2021.
\newblock \href {https://doi.org/10.18653/v1/2021.acl-long.107} {Towards
  argument mining for social good: A survey}.
\newblock In \emph{Proceedings of the 59th Annual Meeting of the Association
  for Computational Linguistics and the 11th International Joint Conference on
  Natural Language Processing (Volume 1: Long Papers)}, pages 1338--1352,
  Online. Association for Computational Linguistics.

\bibitem[{Wang et~al.(2023)Wang, Ivison, Dasigi, Hessel, Khot, Chandu, Wadden,
  MacMillan, Smith, Beltagy, and Hajishirzi}]{wang2024far}
Yizhong Wang, Hamish Ivison, Pradeep Dasigi, Jack Hessel, Tushar Khot, Khyathi
  Chandu, David Wadden, Kelsey MacMillan, Noah~A Smith, Iz~Beltagy, and
  Hannaneh Hajishirzi. 2023.
\newblock \href
  {https://proceedings.neurips.cc/paper_files/paper/2023/file/ec6413875e4ab08d7bc4d8e225263398-Paper-Datasets_and_Benchmarks.pdf}
  {How far can camels go? exploring the state of instruction tuning on open
  resources}.
\newblock In \emph{Advances in Neural Information Processing Systems},
  volume~36, pages 74764--74786. Curran Associates, Inc.

\bibitem[{Wang et~al.(2021)Wang, Choi, Xu, and Yang}]{wang2021putting}
Zijie~J. Wang, Dongjin Choi, Shenyu Xu, and Diyi Yang. 2021.
\newblock \href {https://aclanthology.org/2021.hcinlp-1.8} {Putting humans in
  the natural language processing loop: A survey}.
\newblock In \emph{Proceedings of the First Workshop on Bridging
  Human{--}Computer Interaction and Natural Language Processing}, pages 47--52,
  Online. Association for Computational Linguistics.

\bibitem[{Wankhade et~al.(2022)Wankhade, Rao, and
  Kulkarni}]{wankhade2022survey}
Mayur Wankhade, Annavarapu Chandra~Sekhara Rao, and Chaitanya Kulkarni. 2022.
\newblock \href {https://doi.org/10.1007/s10462-022-10144-1} {A survey on
  sentiment analysis methods, applications, and challenges}.
\newblock \emph{Artificial Intelligence Review}, 55(7):5731--5780.

\bibitem[{Waterschoot et~al.(2022)Waterschoot, van~den Hemel, and van~den
  Bosch}]{waterschoot2022detecting}
Cedric Waterschoot, Ernst van~den Hemel, and Antal van~den Bosch. 2022.
\newblock \href {https://aclanthology.org/2022.coling-1.583} {Detecting
  minority arguments for mutual understanding: A moderation tool for the online
  climate change debate}.
\newblock In \emph{Proceedings of the 29th International Conference on
  Computational Linguistics}, pages 6715--6725, Gyeongju, Republic of Korea.
  International Committee on Computational Linguistics.

\bibitem[{Wei et~al.(2022)Wei, Tay, Bommasani, Raffel, Zoph, Borgeaud,
  Yogatama, Bosma, Zhou, Metzler, Chi, Hashimoto, Vinyals, Liang, Dean, and
  Fedus}]{wei2022emergent}
Jason Wei, Yi~Tay, Rishi Bommasani, Colin Raffel, Barret Zoph, Sebastian
  Borgeaud, Dani Yogatama, Maarten Bosma, Denny Zhou, Donald Metzler, Ed~H.
  Chi, Tatsunori Hashimoto, Oriol Vinyals, Percy Liang, Jeff Dean, and William
  Fedus. 2022.
\newblock \href {https://openreview.net/forum?id=yzkSU5zdwD} {Emergent
  abilities of large language models}.
\newblock \emph{Transactions on Machine Learning Research}.
\newblock Survey Certification.

\bibitem[{Weld et~al.(2022)Weld, Zhang, and Althoff}]{weld2022makes}
Galen Weld, Amy~X. Zhang, and Tim Althoff. 2022.
\newblock \href {https://doi.org/10.1609/icwsm.v16i1.19363} {What makes online
  communities ‘better’? measuring values, consensus, and conflict across
  thousands of subreddits}.
\newblock \emph{Proceedings of the International AAAI Conference on Web and
  Social Media}, 16(1):1121--1132.

\bibitem[{Wiebe et~al.(2005)Wiebe, Wilson, and Cardie}]{wiebe2005annotating}
Janyce Wiebe, Theresa Wilson, and Claire Cardie. 2005.
\newblock \href {https://doi.org/10.1007/s10579-005-7880-9} {Annotating
  expressions of opinions and emotions in language}.
\newblock \emph{Language resources and evaluation}, 39:165--210.

\bibitem[{Wu et~al.(2020)Wu, Wu, Qi, and Huang}]{wu2020sentirec}
Chuhan Wu, Fangzhao Wu, Tao Qi, and Yongfeng Huang. 2020.
\newblock \href {https://aclanthology.org/2020.aacl-main.6} {{S}enti{R}ec:
  Sentiment diversity-aware neural news recommendation}.
\newblock In \emph{Proceedings of the 1st Conference of the Asia-Pacific
  Chapter of the Association for Computational Linguistics and the 10th
  International Joint Conference on Natural Language Processing}, pages 44--53,
  Suzhou, China. Association for Computational Linguistics.

\bibitem[{Xia et~al.(2020)Xia, Zhu, Lu, Zhang, and Gu}]{xia2020exploring}
Yan Xia, Haiyi Zhu, Tun Lu, Peng Zhang, and Ning Gu. 2020.
\newblock \href {https://doi.org/10.1145/3415179} {Exploring antecedents and
  consequences of toxicity in online discussions: A case study on reddit}.
\newblock \emph{Proc. ACM Hum.-Comput. Interact.}, 4(CSCW2).

\bibitem[{Xiong and Liu(2014)}]{xiong2014opinion}
Fei Xiong and Yun Liu. 2014.
\newblock \href {https://doi.org/10.1063/1.4866011} {{Opinion formation on
  social media: An empirical approach}}.
\newblock \emph{Chaos: An Interdisciplinary Journal of Nonlinear Science},
  24(1):013130.

\bibitem[{Xu et~al.(2023)Xu, Walder, and Xu}]{xu2021humanly}
Qiongkai Xu, Christian Walder, and Chenchen Xu. 2023.
\newblock \href {https://openreview.net/forum?id=X5ZMzRYqUjB} {Humanly
  certifying superhuman classifiers}.
\newblock In \emph{The Eleventh International Conference on Learning
  Representations}.

\end{thebibliography}

\end{document}